\pdfoutput=1
\documentclass[11pt]{article}

\usepackage{ACL2023}

\usepackage{times}
\usepackage{latexsym}
\usepackage{multirow}
\usepackage{graphicx}
\usepackage{float}
\usepackage{subfigure}
\usepackage{enumitem}
\usepackage{booktabs}
\usepackage{pifont}
\usepackage{caption}
\usepackage{xcolor,colortbl}
\usepackage{color}
\usepackage[utf8]{inputenc}
\usepackage{tcolorbox}
\usepackage[T1]{fontenc}
\usepackage{amsmath}
\usepackage{amssymb}
\usepackage{stfloats}
\usepackage[utf8]{inputenc}
\usepackage{amsbsy}
\usepackage{microtype}
\usepackage{inconsolata}

\newtcolorbox{conversationbox}{
    colback=white,
    colframe=black,
    sharp corners,
    boxrule=1pt,
    width=0.47\textwidth
}

\title{Enhancing Large Language Model Induced Task-Oriented Dialogue Systems Through Look-Forward Motivated Goals}

\author{First Author \\
  Affiliation / Address line 1 \\
  Affiliation / Address line 2 \\
  Affiliation / Address line 3 \\
  \texttt{email@domain} \\\And
  Second Author \\
  Affiliation / Address line 1 \\
  Affiliation / Address line 2 \\
  Affiliation / Address line 3 \\
  \texttt{email@domain} \\}

\author{Zhiyuan Hu$^{1}$ \quad Yue Feng$^{2}$ \quad Yang Deng$^{1}$ \quad Zekun Li$^{3}$ \\
\quad \textbf{See-Kiong Ng}$^{1}$  \quad \textbf{Anh Tuan Luu}$^{4}$\thanks{\quad Equal Advising} \quad \textbf{Bryan Hooi}$^{1}$\footnotemark[1] \\
$^{1}$National University of Singapore  \ 
$^{2}$University College London  \\
$^{3}$University of California, Santa Barbara 
$^{4}$Nangyang Techonology Unversity \\
        \texttt{zhiyuan\_hu@u.nus.edu}, \space
         \texttt{yue.feng.20@ucl.ac.uk}, \space 
         \texttt{ydeng@nus.edu.sg}, \space
         \texttt{zekunli@cs.ucsb.edu} \\\space
         \texttt{seekiong@nus.edu.sg} \space
         \texttt{anhtuan.luu@ntu.edu.sg} , \space \texttt{bhooi@comp.nus.edu.sg}}

\begin{document}
\maketitle
\begin{abstract}

Recently, the development of large language models (LLMs) has been significantly enhanced the question answering and dialogue generation, and makes them become increasingly popular in current practical scenarios. While unlike the general dialogue system which emphasizes the semantic performance, the task-oriented dialogue (ToD) systems aim to achieve the dialogue goal \textbf{efficiently} and \textbf{successfully} in multiple turns. Unfortunately, existing LLM-induced ToD systems lack the direct reward toward the final goal and do not take account of the dialogue proactivity that can strengthen the dialogue efficiency. To fill these gaps, we introduce the \textbf{ProToD} (Proactively Goal-Driven LLM-Induced ToD) approach, which anticipates the future dialogue actions and incorporates the goal-oriented reward signal to enhance ToD systems. Additionally, we present a novel evaluation method that assesses ToD systems based on goal-driven dialogue simulations. This method allows us to gauge user satisfaction, system efficiency and successful rate while overcoming the limitations of current Information and Success metrics. Empirical experiments conducted on the MultiWoZ 2.1 dataset demonstrate that our model can achieve superior performance using only 10\% of the data compared to previous end-to-end fully supervised models. This improvement is accompanied by enhanced user satisfaction and efficiency. \footnote{Codes and Prompt are released to \url{https://github.com/zhiyuanhubj/ProToD}}

\end{abstract}

\section{Introduction}

A task-oriented dialogue system is designed to assist users in achieving specific objectives. Its primary focus is on comprehending user needs and generating appropriate responses. The success rate is a pivotal metric in evaluating the effectiveness of a ToD system. A higher success rate indicates that the system is adept at meeting user requirements. Additionally, efficiency is gauged by the number of turns in a conversation. Fewer turns signify greater efficiency, underscoring the need for the system to be proactive. Drawing from the concept of proactivity in organizational behaviors \cite{grant2008dynamics} and standard dictionary definitions \cite{english1976oxford}, the proactivity of conversational agents can be characterized as their ability to steer or control a dialogue. This is achieved by taking the initiative and foreseeing potential impacts on themselves or users. In essence, the ultimate success of a ToD system mainly lies in proactive nature and capacity to effectively and efficiently address user needs.

Current research focuses on guiding LLMs to produce relevant responses using task-specific instructions and few examples. \cite{li2023guiding} introduced a method where a small model provides directional prompts for each query. LLMs use these prompts and previous dialogues to respond. This optimizes LLMs in ToDs by adjusting a minor policy model, which can be refined using supervised learning and reinforcement based on BLUE score rewards. Additionally, \cite{hu2023unlocking} presented a framework that uses LLMs as user simulators to enhance task-oriented dialogue models.

However, these methods consistently ignore to make the ToD system more successful and proactive. Current LLM-based ToD systems rely on BLUE score-based or user simulation based rewards, which primarily measure the similarity between generated responses and ground truth or the user satisfaction scores. These approaches may not fully capture the essence of goal-driven dialogue. Moreover, LLMs tend to produce more flexible and longer responses compared to end-to-end models, leading to lower reward values even the wrong reward optimization direction. Therefore, a new reward mechanism that focuses on goal-driven behavior to guides LLM for generating response should be worth exploring. Moreover, since a single dialogue context can often have multiple valid responses, generating a desired response solely based on historical information is a complex task. If the chatbot can anticipate what the user is likely to discuss next after receiving its response, it can provide a response that smoothly connects the past and future elements of the conversation.

%While, such methods still exist two serious problems. Firstly, since one dialogue query can often be appropriately answered by multiple distinct responses. generating a desired response solely based on the historical information is not easy. If the chatbot can be told in advance what the user would probably talk about (i.e., the dialogue future) after receiving its response, it only needs to provide a response that can smoothly connect the history and the future.  Secondly, completing the dialogue goal is the most crucial aspect of ToD system. While the current DSP work is optimized based on the BLUE score based reward which only consider the similarity of generated response and ground truth. Additionally, current LLMs' responses are more flexible and longer than End-to-end models' generation results, which led to the lower reward value even the wrong optimization direction. Hence, the new reward based on the goal-driven to drive guiding LLM cues generation should be proposed.

\begin{figure}
    \centering
    \includegraphics[width=0.5\textwidth]{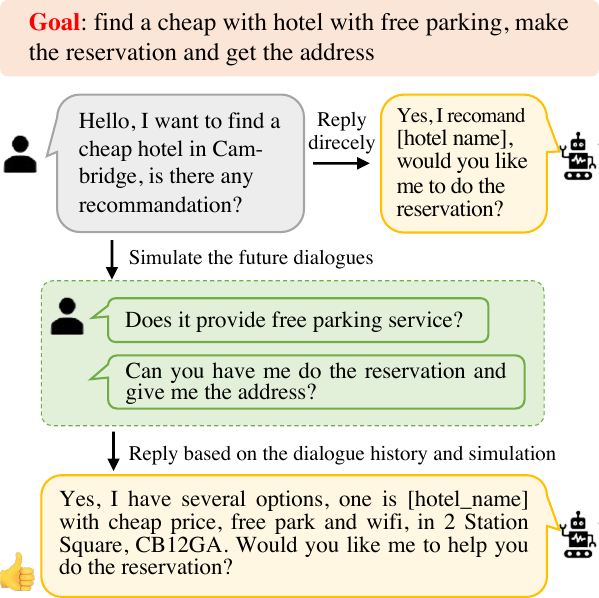}
    \caption{An example of dialogue response generation when considering future dialogues and dialogue goal}
    \label{fig:explanation example}
\end{figure}

%To address the above challenges, we propose the \textbf{M}otivation of \textbf{G}oal (\textbf{LFMG}) method including \textbf{Anticipating Future Actions} and \textbf{Goal-oriented Reward}. As illustrated in \ref{fig:explanation example}, It will be beneficial for the ToD system to foresee the users' demands and prepare more diverse and comprehensive response when the model anticipates the future utterance actions rather than only predicting next utterance action. Hence, the ToD system will be easier to satisfy the users' requirements, enhance the goal successful rate and improve the dialogue efficiency. Additionally, compared with utilizing BLUE or user simulation scores as the reward, leveraging the goal completion degree as the reward to optimize LLM-induced ToD system is a more natural and reasonable method. We present a new reward calculation method by taking into account how many sub-goals each system reponse fullfill to replace the previous semantic similarity based or user feedback based reward function.

To address the aforementioned challenges, we propose the \textbf{ProToD} (Proactively Goal-Driven LLM-Induced ToD) method, which encompasses two crucial components: \textbf{Anticipating Future Actions} and \textbf{Goal-oriented Reward}. As depicted in Figure~\ref{fig:explanation example}, the ability of a ToD system to anticipate users' future demands and prepare a more diverse and comprehensive response can be highly advantageous. This approach goes beyond merely predicting the next user utterance action, making it easier for the ToD system to meet users' requirements, elevate the success rate of achieving goals, and enhance overall dialogue efficiency. Furthermore, in contrast to using metrics like BLUE scores or user simulation scores as the basis for reward calculation, opting for the degree of goal completion as the reward offers a more natural and rational approach to optimizing LLM-induced ToD systems. In this context, we introduce a novel reward calculation method that considers the extent to which each system response fulfills sub-goals, replacing the previous reliance on semantic similarity or user feedback-based reward functions. 

Moreover, the current metrics, such as ``Inform'' and ``Success'', which rely on fixed ground-truth values, lack the flexibility to accurately gauge effectiveness and success rates. In addition, prior research \cite{wu2023using} has highlighted that when responses are consistently fixed as "predefined responses" in every turn, it leads to superior outcome. This phenomena indicates these metrics can not be a suitable one. As a result, we propose a novel evaluation method that employs GPT-4 \cite{openai2023gpt4} as the user simulator. In this approach, users are required to adhere to predefined goals when interacting with the ToD system. The extent to which these conversations successfully achieve their goals and the number of turns required are used to measure both success and efficiency.

To summarize, our contributions in this work are these three perspectives:
\begin{itemize}
    \item We propose the \textbf{ProToD} (Proactively Goal-Driven LLM-Induced ToD) approach which anticipates future dialogue actions and integrates a goal-oriented reward signal, enhancing the efficiency and success of ToD systems.
    \item To better and flexibly evaluate the efficiency and successful rate of LLM-induced ToD system, we introduce the goal-driven user simulation based on GPT-4 to assess the performance ToD system.
    \item We conduct the comprehensive experiments including automatic metrics evaluation, user simulator based assessment, case study and human evaluation, which fully validate the effectiveness of our model.
\end{itemize}

\section{Related Work}
TOD systems facilitate tasks like hotel bookings or restaurant reservations. Previous end-to-end models \cite{he2022galaxy, lee2021improving, sun2023mars, wu2023using} generate responses using only the dialogue context, while policy optimization methods \citet{wang2020modelling, wang2020multi} uses ground-truth dialogue states. \citet{lubis2020lava} and \citet{ lee2021improving} incorporate both text information and dialog states to generate the dialogue response. Additionally, reinforcement learning methods \cite{wu2023diacttod, bang2023task, feng2023fantastic} have also been validated for improvement of TOD systems.

In terms of the LLM-based ToD research, \citet{madotto2020language} assess the few-shot capability of language models in Natural Language Understanding, Dialogue State Tracking, Dialogue Policy and Natural Language Generation tasks. \citet{hudevcek2023llms} evaluate Instruction-finetuned LLMs' ability to complete multi-turn tasks and interact with external databases in the context of established task-oriented dialogue benchmarks. \citet{snell2022context} formulate goal-oriented dialogue as a partially observed Markov decision process, interpreting the language model as a representation of both the dynamics and the policy. Recently, \citet{hu2023unlocking} propose a new framework to leverage LLM as the user simulator and utilize the feedback of this simulation to optimize the ToD model. \citet{li2023guiding} introduce a novel prompting framework called Directional Stimulus Prompting for guiding black-box LLMs toward desired output, which employ a small tunable policy model to generate the hint to guide the LLMs.

\section{Methodology}

\begin{figure*}
    \centering
    \includegraphics[width=1\textwidth]{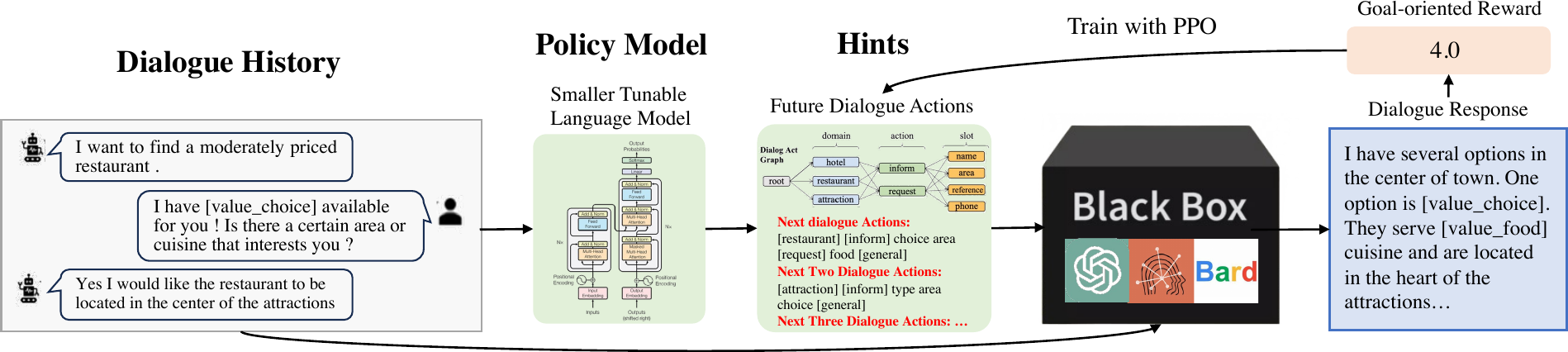}
    \caption{Model Architecture. The policy model is trained for generate future dialogues actions to induce LLMs in specifc tasks by supervised finetuning and PPO based reinforcement learning.}
    \label{fig:enter-label}
\end{figure*}

\subsection{Overview}

For dialogue response generation, we consider an input dialogue history space denoted as $\mathbf{X}$, a data distribution represented by $\mathcal{D}$ over $\mathbf{X}$, and a response output space referred to as $\mathbf{Y}$. LLMs have shown remarkable capabilities in generating responses by incorporating instructions that describe the task, a few demonstration examples, and the input dialogue history $\mathbf{x}$ within the prompt. However, there are challenges when it comes to steering LLMs towards desired outputs, particularly for achieving fine-grained, query-specific behaviors. In the context of ToD, different dialogue systems need to respond to user queries across various domains using actions such as informing, requesting, confirming, and providing domain-specific slot values. In such scenarios, solely relying on task-specific instructions and a handful of examples may not consistently yield satisfactory and relevant responses. Additionally, dealing with long-term memory and maintaining efficiency poses further challenges in LLM-based ToD systems.

To address these issues, we propose the incorporation of future dialogue action hints denoted as $\boldsymbol{z}$ into the prompt, inspired by the Directional Stimulus Prompting (DSP) approach \cite{li2023guiding}. These hints serve as guidance for achieving the desired response. For each input query, we generate these hints using a small, adaptable policy language model, $p_{POL}(\boldsymbol{z}|\boldsymbol{x})$. Subsequently, we combine the generated hint, $\boldsymbol{z}$, with the original dialogue history, $\boldsymbol{x}$, to construct the prompt that guides the LLM towards generating its output, represented as $p_{LLM}(\boldsymbol{y}|\boldsymbol{x},\boldsymbol{z})$.

\subsection{Anticipating Future Dialogues Actions}

To train the policy model that generates guided future response actions for LLMs, we first perform supervised
fine-tuning on a pre-trained LM such as T5 on a small collection of labeled data (1\% or 10\%). 

%For better enable LLMs to generate the task-specific response, we use following turns' dialogue acts as the hint $\boldsymbol{z}$ that indicate the underlying meaning of the desired system response. The resulting dataset $\mathcal{D} = \{\boldsymbol{x}, \boldsymbol{z}\}$ consists of dialogue history-future actions pairs. Specificly, for the dialogue history with $n$ turns $\mathbf{x} = (x_1, x_2, x_3 \ldots x_n)$ and each response turn corresponding action $\mathbf{a} = (a_1, a_2, a_3 \ldots a_n)$, The $i$-th turn's response future action can be formulated as $z_i = a_i, a_{i+1} \ldots a_n $. We then fine-tune the policy model by maximizing the log-likelihood:

To enhance the ability of LLMs to generate task-specific responses, we employ the anticipated future dialogue actions, spanning from the current turn until the end of the conversation, as contextual cues for guiding the LLM in generating responses to the queries from the current user turn. These cues are denoted as denoted as $\boldsymbol{z}$, which convey the intended meaning of the desired system response. The resulting dataset, denoted as $\mathcal{D} = {(\boldsymbol{x}, \boldsymbol{z}})$, comprises pairs of dialogue histories and future action sequences. Specifically, given a dialogue history with $n$ turns, represented as $\mathbf{x} = (x_1, x_2, x_3, \ldots, x_n)$, and corresponding actions for each response turn, denoted as $\mathbf{a} = (a_1, a_2, a_3, \ldots, a_n)$, we formulate the future action for the $i$-th turn as $z_i = (a_i, a_{i+1}, \ldots, a_n)$. Subsequently, we fine-tune the policy model by maximizing the log-likelihood through the following objective:
\begin{equation}
   \mathcal{L}_{\mathrm{POL}}=-\mathbb{E}_{\left(\boldsymbol{x}, \boldsymbol{z}\right) \sim \mathcal{D}} \log p_{\mathrm{TOD}}\left(\boldsymbol{z} \mid \boldsymbol{x}\right) 
\end{equation}

This framework enables our model to generate responses that align with the underlying dialogue acts, resulting in more contextually appropriate and task-specific outputs for the current user query. For better modify this future dialogue actions hints toward the desired outputs with high successful rate for dialogue goal, we continue to incorporate reinforcement learning (RL) to further fine-tune the policy model based on the goal-oriented reward. The detailed content is elaborated below.

\subsection{Goal-oriented Reward}

Our objective is to guide the generation of the LLM towards our desired target by optimizing an alignment measure denoted as $\mathcal{R}$. In this context, we define the competency level of the dialogue goal as our reward.

In each dialogue, the overall goal can be subdivided into several predefined sub-goals, denoted as $\mathbf{g} = (g_1, g_2, g_3, \ldots, g_n)$. The success of each sub-goal can be measured by assessing whether the ToD system provides the corresponding information. For instance, in the MultiWoZ dataset, these sub-goals encompass tasks such as supplying a reference ID, disclosing a phone number, furnishing an address, and so forth. Consequently, for each system response turn, we can calculate a turn-level goal reward that quantifies how many sub-goals it accomplishes. This can be mathematically formulated as follows:
\begin{equation}
    r_i = \lambda \sum_{j=0}^{n} g_j
\end{equation}
where $r_i$ represents the reward for the $i$-th turn, and $\lambda$ is a hyperparameter to scale the reward. The value of each $g_j$ is determined by whether the ToD system provides the corresponding information: if it does, $g_j$ is assigned a value of 1; otherwise, it is set to 0.

Meanwhile, given that the parameters of the black-box LLM are neither accessible nor adjustable, we resort to enhancing the policy model's optimization. This involves generating future dialogue actions as hints, which in turn direct the LLMs' generation process towards the maximization of our defined objective.
\begin{equation}
\mathcal{R}_{LLM}(\mathbf{x},\mathbf{z}) = \mathcal{R}(\mathbf{x},\mathbf{y})
\end{equation}
\begin{equation}
    \mathbf{y} \sim p_{LLM}(|(\mathbf{x},\mathbf{y}))
\end{equation}

However, the optimization approach described above poses an intractable problem for the policy model. In order to tackle this challenge, we reframe the optimization of the policy model as a reinforcement learning (RL) problem and leverage the proximal policy optimization (PPO) algorithm \cite{schulman2017proximal}.

We use the policy model to initialize a policy network
$\pi_0 = p_{POL}$ and then update $\pi$ using PPO.

The process through which the policy model generates a sequence of future actions $\boldsymbol{z}$ can be viewed as an interaction in the context of RL, defined by the tuple $⟨\mathbf{S}, \mathbf{A}, \mathbf{r}, \mathbf{P}⟩$. Here, $\mathbf{S}$ represents the state space, $\mathbf{A}$ denotes the action space, $\mathbf{r}$ corresponds to the reward function, and $\mathbf{P}$ signifies the state-transition probability. In each interaction with the environment, the agent selects an action (token) based on the probability distribution defined by the current policy network $\pi(\mathbf{z}|\mathbf{x},\mathbf{z}<\mathbf{t})$. The interaction process concludes when an end-of-sequence token is chosen, resulting in the generation of the entire sequence of future dialogue actions. The policy network $\pi$ can be improved through fine-tuning, aiming to optimize the reward $\mathbf{r}$ associated with the RL framework.

% formulation here

To avoid the policy network $\pi$ deviate from the initial policy model $p_{POL}$ too far, we also introduce the KL-divergence penalty into the current reward function. Therefore, the final rewrad formuls will be:
\begin{equation}
r(\boldsymbol{x}, \boldsymbol{y})=LLM(\boldsymbol{x}, \boldsymbol{y})-\beta \log \frac{\pi(\boldsymbol{y} \mid \boldsymbol{x})}{p_{\mathrm{TOD}}(\boldsymbol{y} \mid \boldsymbol{x})}    
\end{equation}

\vspace{-3mm}
\begin{table*}[ht] 
\centering
\begin{tabular}{l|c|cc}
\toprule
    \multirow{2}{*}{\bf Model}  &\bf Training & \multicolumn{2}{c}{\bf MultiWoZ 2.1}  \\
    &\textbf{Data} & Inform & Success   \\
\midrule
    Standard Prompting  &- &72.8 &44.2    \\
\midrule
    DSP \cite{li2023guiding}  &1\% &87.3 &78.7   \\
    ProToD- &1\% &90.4    &80.1      \\ 
    ProToD &1\%   &94.3 &82.7  \\
\midrule    
    DSP \cite{li2023guiding}  &10\%  & 95.0 &84.0     \\ 
    ProToD- &10\% &95.3 &85.0   \\
    ProToD	&10\% &\textbf{96.2} &85.8  \\ 
\midrule
    SimpleTOD \cite{hosseini2020simple} &100\% &85.0 &70.5 \\
    PPTOD \cite{su2021multi} &100\% &87.1 &79.1 \\
    UBAR \cite{yang2021ubar} &100\% &95.7 &81.8 \\
    GALAXY \cite{he2022galaxy}  &100\% & 95.3 &\textbf{86.2}  \\ 
\bottomrule

\end{tabular}
\caption{Performance of system response generation.}
\vspace{-5mm}
\label{main result}
\end{table*}

\subsection{Goal-driven User Simulation Assessment}

As highlighted by \cite{wu2023using} in their study on current Inform and Success metrics implementation, there are inherent issues. These two metrics evaluate cumulative belief states for placeholders and the presence of references in responses to calculate scores. Consequently, a model that generates more placeholders can misleadingly appear to perform better. When a fixed response such as "[value\_name] [value\_phone] [value\_address] [value\_postcode] [value\_reference] [value\_id]" is consistently used for every turn during evaluation with the standardized evaluation script, it yields state-of-the-art results in terms of Inform and Success scores when compared to baseline models.

Therefore, these problems further necessitate the new method to assess the performance of LLM-based ToD system. Given a dialogue goal $\boldsymbol{g}$, we design the feasible prompt to enable GPT-4 to act as the user to propose the user queries $x^*=\{x_1^*, x_2^*, x_3^* \ldots x_n^* \}$, then the LLM-based ToD will generate the responses $y^*=\{y_1^*, y_2^*, y_3^* \ldots y_n^* \}$ between the turn-by-turn interaction. In these simulations, the efficiency can be calculated as the average value of turns' number. Due to the strong understanding of GPT-4, the success will be assess by GPT-4 again according to the dialogue goal and this simulation dialogue.

%\vspace{-3mm}
\section{Experiments}

\subsection{Dataset, Evaluation Metrics and Implementation Details}

\textbf{MultiWoZ 2.1} \cite{eric-etal-2020-multiwoz} is the improved version of MultiWOZ 2.0 \cite{budzianowski-etal-2018-multiwoz} which is a released multi-domain dialogue dataset spanning 7 distinct domains and containing over 10,000 dialogues. Moreover, MultiWOZ 2.1 also includes user dialogue acts as well as multiple slot descriptions per dialogue state slot.

\noindent \textbf{Inform and Success} are the metrics related to dialogue task completion - whether the system provides an appropriate entity (Inform rate) and then answered all the requested attributes (Success rate)

\noindent \textbf{Implementation}: We employ T5 \cite{raffel2020exploring} as the policy model and leverage GPT-3.5-turbo \citep{OpenAIChatGPT} as the specific LLM here.

\subsection{Prior Performances}

\textbf{Standard Prompting}: design the instruction to let LLMs to reply the previous dialogue history

\noindent \textbf{DSP} \cite{li2023guiding}: In this approach called Directional Stimulus Prompting (DSP), a new element called "directional stimulus" is introduced into the prompt to provide more precise guidance for LLMs. This stimulus acts as cues, like keywords, to guide LLMs in generating desired outputs. A small tunable model, such as T5, is used to create this stimulus for each query, allowing optimization of LLMs through a smaller policy model. This policy model is trained through supervised fine-tuning with labeled data and reinforcement learning using rewards, aiming to align LLM behavior with desired outcomes.

\noindent \textbf{SimpleTOD} \cite{hosseini2020simple} adopts a unified approach, treating all these sub-tasks as a single sequence prediction problem, leveraging pre-trained, open-domain, causal language models like GPT-2 as its base model

\noindent \textbf{PPTOD} \cite{su2021multi} introduces a novel multi-task pre-training strategy for learning TOD skills from diverse dialog datasets

\noindent \textbf{UBAR} \cite{yang2021ubar} is a ToD system that excels in modeling entire dialog sessions by fine-tuning GPT-2 on sequences encompassing user input, belief states, system actions, and responses. UBAR's strengths lie in session-level training and dialog context, making it effective as an end-to-end system, with transferability to new domains.

\noindent \textbf{GALAXY} \cite{he2022galaxy}: This method is the previous end-to-end fully supervised training SOTA model. The Galaxy model is an innovative pre-trained conversational system that effectively acquires dialog strategies by leveraging a combination of limited labeled dialog data and extensive unlabeled dialog datasets through a semi-supervised learning approach. They incorporate a task to predict dialog acts as a means of improving dialog policy during the pre-training phase and utilize a consistency regularization component to enhance the acquired representations with the aid of unlabeled dialog data.

%\subsection{\dy{Implementation Details}}
%\dy{In the current version, it is unclear what is adopted as the policy model. If this subsection is added, 4.1-4.3 would be better to under the section of Experimental Setups, while 4.4-4.6 under a new section of Experimental Results.}

\subsection{ProToD Performance}

As demonstrated by the Table~\ref{main result}, the ProToD model consistently outperforms the DSP model, both when trained on 1\% and 10\% of the data. Notably, even with just 1\% of the data for training the tunable policy model, the ProTOD model exhibits a superior performance with an average improvement of 5.5\% compared to the DSP model. In the case of 10\% training data, our model even surpasses the performance of the 100\% data supervised End-to-End training model (GALAXY) in terms of the Inform metric.

Furthermore, we conducted an ablation study (ProToD\textbf{-}) to assess the effectiveness of future dialogue action anticipation when we retain the previous semantic reward. The experimental results demonstrate that this module alone yields positive effects in both Inform and Success metrics.

\subsection{Goal-driven User Simulation Assessment}

We randomly sample 100 dialogues and employ GPT-4 as the user simulator to conduct the dialogue simulation. Then, calculate the efficiency and let GPT-4 to assess whether this dialogue complete the dialogue goal (success) and how satisfactory (1-5) the user feel. 

As illustrated in Table ~\ref{tab:user assessment results}, ProToD surpasses DSP with a success rate of 69.2\% against DSP's 61.5\%. This indicates ProToD's enhanced capability to achieve dialogue goals. When considering efficiency, ProToD's score of 5.7 outshines DSP's 6.3, suggesting that ProToD requires fewer interactions for task completion. Furthermore, the higher satisfaction score of 4.3 for ProToD, compared to DSP's 4.0, implies that users may find interactions with ProToD more intuitive and satisfying. This holistic improvement highlights ProToD's potential as a superior dialogue system.

\begin{table}[]
    \centering
    \begin{tabular}{c|ccc}
    \toprule
         Method &SU$\uparrow$  &EF$\downarrow$ & SA$\uparrow$ \\
    \midrule
         DSP \cite{li2023guiding} & 61.5 & 6.3 &4.0 \\
         ProToD &69.2 & 5.7 &4.3 \\
    \bottomrule
    \end{tabular}
    \caption{Evaluation performance of dialogue response generation through simulation by GPT-4, where SU, EF and SA denote as Successful Rate, Efficiency and Satisfaction respectively}
    \label{tab:user assessment results}
    \vspace{-5mm}
\end{table}

\subsection{Case Study}

In our study, we employ a case study approach to further bolster our motivation. This case study comprises four key components: the conversation goal, history, DSP model response, and ProToD model response. When we examine the ProToD response in comparison to the DSP response, it becomes evident that the ProToD model offers two notable advantages. Firstly, the ProToD response excels in providing current options and strives to meet the user's objectives as efficiently as possible, rather than simply asking for the type of food, thereby significantly enhancing dialogue efficiency. Additionally, for each option presented, the ProToD-generated response includes comprehensive information encompassing cuisine types and location clarifications. Secondly, the ProToD response adeptly addresses the user's intent confirmation by providing a phone number when required, potentially increasing the success rate of the dialogue. These two key strengths can further support the enhancement of Inform and Success metrics within the MultiWoZ 2.1 dataset.

\begin{conversationbox}

\small{\textbf{Conversation Goal}:

You are planning your trip in Cambridge. You are looking for a \textbf{restaurant}. The restaurant should be in the \textbf{centre} and should be in the \textbf{moderate} price range. Make sure you get \textbf{food type} and \textbf{postcode}. You are also looking for \textbf{places to go} in town. The attraction should be in the type of \textbf{mutliple sports} and should be \textbf{in the same area as the restaurant}. If there is no such attraction, how about one that is in \textbf{the type of architecture}. Make sure you get \textbf{phone number}, \textbf{entrance fee}, and \textbf{address}} \\

\textbf{Conversation History}:

User:  I want to find a moderately priced restaurant .

Assistant:  I have [value\_choice] available for you ! Is there a certain area or cuisine that interests you ?

User:  Yes I would like the restaurant to be located in the center of the attractions . \\

\textbf{\textcolor{red}{DSP Response:}} There are [value\_choice] restaurants in the center of the attractions . what type of food are you interested in ? \\

\textbf{\textcolor{red}{ProToD Response:}} I have several options in the center of town. One option is [value\_choice]. They serve [value\_food] cuisine and are located in the heart of the attractions. Another option is [value\_choice]. They have a variety of cuisines and are also centrally located. Would you like me to provide you with their phone? 
\end{conversationbox}

%\section{Human Evaluation}

\section{Conclusion}

In this study, we present the \textbf{ProToD} model, an enhancement of the LLM-induced ToD system that incorporates future dialogue action anticipation and goal-oriented reward motivation. By utilizing future actions as cues to guide LLMs, our model offers more comprehensive responses and enhances the efficiency of dialogues. The integration of goal-oriented rewards further fine-tunes the cues for LLMs, resulting in improved dialogue task completion rates through a reinforcement learning framework. Additionally, we introduce a goal-driven user simulation assessment based on GPT-4, providing a novel perspective to better evaluate dialogue efficiency and user satisfaction levels. Our validation process assesses the effectiveness of ProToD by examining performance enhancements in Inform and Success metrics using the MultiWoZ 2.1 dataset. Furthermore, we present case studies and user simulation assessments that illustrate the improvements in dialogue efficiency and user satisfaction achieved by our model.

\bibliography{custom}
\bibliographystyle{acl_natbib}

\end{document}